\documentclass{article}
\usepackage{ijcai21}
\usepackage{times}
\usepackage{soul}
\usepackage{url}
\usepackage{hyperref}
\usepackage[utf8]{inputenc}
\usepackage[small]{caption}
\usepackage{graphicx}
\usepackage{amsmath}
\usepackage{amsthm}
\usepackage{booktabs}
\usepackage{algorithm}
\usepackage{algorithmic}
\usepackage{xcolor}
\title{Anomaly Mining - Past, Present and Future
\thanks{\em A position paper for IJCAI 2021 Early Career Spotlight Talk}
}

\author{
    Leman Akoglu
    \affiliations
    Heinz College, Carnegie Mellon University\\
    Pittsburgh, PA
    \emails
    lakoglu@andrew.cmu.edu
}

\begin{document}

\maketitle

\begin{abstract}
	
Anomaly mining is an important problem that finds numerous applications in various real world domains such as environmental monitoring, cybersecurity, finance, healthcare and medicine, to name a few.
In this article, I focus on two areas, (1) point-cloud and (2) graph-based anomaly mining. I aim to present a broad view of each area, and discuss classes of main research problems, recent trends and future directions.
 I conclude with key take-aways and overarching open problems.

{\bf Disclaimer.} 
I try to provide an overview of past and recent trends in both areas within 4 pages.
Undoubtedly, these are my personal view of the trends, which can be organized differently. 
For brevity, I omit all technical details and refer to corresponding papers.
Again, due to space limit, it is not possible to include all (even most relevant) references, but a few representative examples.
I aim to keep this as a working document, possibly to be extended based on reader feedback.
 
\end{abstract}


\section{Point-cloud Anomaly Mining}

Point-cloud data consists of points that reside in a feature space, each of which can be seen as a $d$ dimensional vector.
Anomalous points are typically referred to as outliers, and in this section I will adopt this terminology.

Outlier mining has a very large literature, where most attention has been given to outlier detection (OD) under various settings \cite{books/sp/Aggarwal2013}.
There exist a large pool of detectors that are distance-based, density-based, statistical-, cluster-, angle-, and depth-based, among many others \cite{chandola2009anomaly}.
Most detection models assume outliers to be scattered isolate points, while some specifically aim to detect collective outliers that can be seen as micro-clusters \cite{han2012outlier}. Another class of detectors target contextual outliers, which stand out within a specific context \cite{liang2016robust,conf/pkdd/MeghanathPA18}. These can also be seen as conditional outliers \cite{song2007conditional}.
In addition, dynamic/streaming point-cloud OD has been studied at large \cite{conf/ijcai/TanTL11,gupta2013outlier,conf/kdd/ManzoorLA18} as outliers may often arise in settings where data is collected and monitored over time.

In the rest of this section, I discuss some of the trending classes of problems in outlier mining,
organized into four lines of work as 
(1) user-centric OD, 
(2) deep learning based OD, 
(3) automating OD, and
(4) fairness-aware OD.




\subsection{User-centric Outlier Detection}


User-centric outlier mining comprises two related topics: ($i$) explanations, and
($ii$) human-in-the-loop detection (HILD).
Explaining the detected anomalies is crucial for settings in which outliers need to be vetted by human analysts.
The purpose of vetting could be root-cause analysis/troubleshooting or sanity-checking/justification.
An example to the former scenario is when the analyst identifies faults or inefficiencies in a production line or data center through OD and aims to \textit{fix} the issues generating these outliers. 
The aim for the latter scenario is to distinguish statistical outliers from domain-relevant ones, where e.g. in claims auditing, not all outliers are necessarily associated with fraud.
Related, HILD aims to leverage human feedback for
 sieving mere statistical outliers out of domain-relevant ones to eliminate false positives and thereby increase detection rate.
These two problems are intertwined, since explanations could be presented to human analysts for acquiring effective feedback during HILD.

Although the vast body of work on outlier explanations is recent,  the earliest example dates back several decades \cite{conf/vldb/KnorrN99}, which provided what is called ``intensional knowledge'' by identifying minimal subspaces in which outliers stand out. 
Most existing work in this area are discriminative, since explanation proceeds detection that outputs (outlier/inlier) labels,
 and aim to identify subspaces that well-separate the outliers from the inliers \cite{conf/icde/DangANZS14,conf/aaai/KuoD16,liu2017contextual}. While these have focused on providing a separate explanation for each outlier, others aim to provide explanations for \textit{groups} of outliers \cite{journals/datamine/MachaA18,conf/pkdd/GuptaESAF18} with the intent to reduce information overload on the analyst.

On the other hand, interactive OD 
mainly 
aims to leverage the (ground-truth) labels provided by a human-analyst during an auditing process to
maximize the total number of true anomalies
shown within a given auditing budget \cite{conf/icdm/DasWDFE16}. 
In addition to detection precision, others also factor human-effort in the overall objective
\cite{ghani2011interactive,conf/sdm/LambaA19,chai2020human}. 

Some of the remaining challenges in user-centric OD include
causal explanations and novel
paradigms for human-in-the-loop OD.
Specifically, when the purpose is putting the explanations into action, e.g. fixing the production line,
explanations should reflect a causal relation with the outcome, whereas work in this area is currently quite limited \cite{janzing2019causal}.
As for interactive OD, ideas at the intersection of OD and crowdsourcing could further be explored to design effective ways of leveraging human input.

\subsection{\bf Deep Learning for Outlier Detection}

With the advent of deep neural networks (NNs),
deep learning based OD has attracted attention recently. 
The main lines of work include
reconstruction-based techniques (e.g. via autoencoding 
or generative adversarial networks), 
one-class classification based end-to-end training, 
or simply using deep NNs for representation learning that is better suitable for OD.
Some approaches also leverage synthesized pseudo-outlier points or estimate what-is-called pseudo-labels for existing points. I refer to three surveys for extensive details \cite{chalapathy2019deep,journals/corr/abs-2009-11732,pang2021learning}.

Deep NNs have real prowess for many learning problems, especially those in NLP and computer vision. However, a major obstacle to achieving similar ground-breaking performance on OD for such models appears to be  model selection.
Deep NNs exhibit several hyperparameters (e.g. depth, width, learning rate, drop-out rate, weight decay, number of training epochs, etc.). In the face of lack of any labels, it remains a big hurdle to effectively setting them for unsupervised OD tasks.

\subsection{\bf Automating Outlier Detection}

Provided a large body of detection models exists, each with their own assumptions, a key challenge is to determine which model would be most suited for a given new task.
Moreover, most outlier detectors exhibit (hyper)parameters (HPs) to be set by users, e.g. number of nearest neighbors (NNs) $k$ for NN-based methods \cite{conf/sigmod/BreunigKNS00}. 
It is understood that most detectors are quite sensitive to the choice of their HPs \cite{journals/sigkdd/AggarwalS15,Campos2016,goldstein2016comparative}.
This is perhaps especially the case for deep NN based OD models.
As such, the problem of selecting an algorithm (i.e. detection model) as well as its associated HPs is a critical one.

There has been some work on outlier model selection in recent years.
AutoOD  \cite{li2020autood} focuses specifically on a deep autoencoder based detection model and automatically searches for an optimal neural network architecture within a predefined search space. Here they assume the availability of some labeled data that is used for validation.
MetaOD \cite{zhao2020automating} employs meta-learning, transferring knowledge from historical OD tasks in the form of various models' performances to selecting a model for a given new task {\em without requiring any labels}.


At large, automating outlier model selection, especially for unsupervised settings, remains to be a vastly understudied, yet extremely important area.


\subsection{\bf Fairness-aware Outlier Detection}

Fair data mining and OD are close cousins, as it is exactly the goal of OD to spot rare, minority samples in the data. However statistical minorities, including those associated with certain societal minorities (e.g. Hispanic), do not always reflect positive-class membership (e.g. fraud).
Therefore, if care is not taken when employing OD on data involving human subjects, OD can produce 
unjust outcomes. 

The outlier mining community has recently routed attention to fairness-aware detection. 
One of the earliest work \cite{conf/ecai/DavidsonR20} focused on quantifying the fairness of any OD model's output post hoc (i.e., proceeding detection). FairLOF \cite{deepak2020fair} aimed to incorporate fairness specifically to the LOF algorithm \cite{conf/sigmod/BreunigKNS00}.
DeepFairSVDD \cite{zhang2021towards} employed adversarial training, besides the SVDD objective, to obfuscate protected group memberships.
All of these aim to achieve statistical (a.k.a. group or demographic) parity (SP) as the notion of fairness.
In our recent work \cite{shekhar2020fairod}, we discussed potential sources and implications of bias in OD outcomes, and
designed FairOD targeting additional fairness criteria for OD besides SP, such as treatment parity and equality of opportunity.

One of the key challenges in fair OD is the absence of any ground-truth outlier labels.
Moreover, which notions and measures of fairness are suitable for OD, or the compatibility thereof, is not well understood.


\section{Graph-based Anomaly Mining}

Graphs are powerful representations of relational data in which entities (nodes) are linked through relationships (edges).
They may exhibit various properties;
nodes can be associated with labels or (multi-) attributes, edges can be directed, weighted, and have multiplicities, types, timestamps, etc. 
For example, consider an accounting graph database in which each graph depicts an itemized transaction; where nodes correspond to accounts, node labels depict account types (cash, revenue, etc.), edge multiplicities and weights respectively reflect the number of items  and dollar amount involved, 
edge directions depict debit/credit flow, and finally graph-level attributes denote auxiliary information about the entire graph, such as the time, approver, etc. 
Alternately, consider a single bipartite graph depicting user--product review relations, in which both users and products exhibit multi-attributes (resp., e.g. username, location, etc. and brand, price, etc.), and edges have timestamps and are associated with a rating, and review text.
These examples are to suggest that graphs in the real world can be quite complex. 

The representational complexity (or expressiveness) of real world graphs has driven two key  
challenges for graph anomaly detection (GAD): (1) defining what constitutes a graph anomaly, and (2) designing general-purpose GAD solutions that can seamlessly handle graphs with any (subset of) properties without much/any modification. 
In the following, I discuss prevalent trends on graph anomaly detection problems, as well as  graph neural network based techniques toward automated representation learning for complex graphs. 



\subsection{Defining Graph Anomalies}

Graph anomalies can be organized into three main classes of problems: ($i$) node/edge-level,
($ii$) subgraph-level, and ($iii$) graph-level anomaly detection.
Within each class, the definition of anomaly varies, mainly driven by two factors; the application and the available graph properties. 
I refer to our survey \cite{akoglu2015graph} for an outline of detection problems for static and dynamic, as well as plain and attributed graphs. I aim to give an overview of prevalent definitions and key approaches as follows.

\textbf{Node/edge-level} anomaly detection aims to identify rare or unusual nodes in a single input graph.
{\bf \em Feature-based} approaches \cite{conf/pakdd/AkogluMF10} ``flatten'' the graph into individual node vectors, effectively defining anomalous nodes as point-cloud outliers in this feature space.
{\bf \em Proximity-based} approaches define anomalies in relation to others, where the assumption is that anomalous nodes  are associated with/close-by to other anomalous nodes. This is typically applied to fraud detection problems, where fraudulent users/accounts/etc. can be found nearby in the graph \cite{conf/kdd/RayanaA15}. 
Note that in this case, the definition is motivated by the application.
Another category of approaches is {\bf \em community-based}, which define anomalies as ``misfits'' with respect to the community structure. Those could be anomalous (bridge) edges between communities \cite{conf/sdm/TongL11,conf/icde/AggarwalZY11}, or community outliers that are within well-structured communities and yet, do not exhibit similar node attributes to others in the community \cite{gao2010community,perozzi2014focused} (akin to contextual outliers in point-cloud OD).
Note that in the latter case, the definition is informed by the availability of node attributes.

{\textbf{Subgraph-level}} anomaly detection aims to identify unusual \textit{groups} of nodes in a single input graph. These are akin to collective outliers in point-cloud OD.
{\bf \em Density-based} approaches define subgraph anomalies as those with significantly higher density than the overall graph \cite{hooi2016fraudar}. Provided edge timestamps, definition is also extended to those with high-density and also temporally-synchronized edges \cite{beutel2013copycatch}.
{\bf \em Community-based} approaches leverage node attributes to define anomalous subgraphs as those communities with poor structural connectivity and/or incoherent attribute values \cite{perozzi2016scalable}. 

Finally, \textbf{graph-level} anomaly detection aims to identify unusual graphs within a (unordered) \textit{set} or a (often temporal) \textit{sequence} of graphs.
Most dominant category of approaches is {\bf \em feature/similarity-based}, which either
``flatten'' the entire graph into a single feature vector, or quantify pairwise similarity between graphs through an appropriately designed graph kernel \cite{manzoor2016fast}. 
 These effectively treat anomalous graphs as point-cloud outliers in an explicit or implicit feature space.
{\bf \em Subgraph-based} approaches specifically define anomalous graphs as those with (the (dis)appearance of)  dense subgraphs \cite{eswaran2018spotlight} 
or with (the emergence of) ``hotspots'' \cite{hooi2018changedar}, depending on whether anomalies are sought within a static set of graphs or time-evolving series of graphs.
While the former is defined solely topologically based on edge density, the latter definition is driven by node-level signals (i.e. continuous attributes) where hotspots are defined as those groups of close-by nodes in the graph whose signal values are or have become abnormal.
Another classic type of approaches is {\bf \em compression-based}, which define anomalous graphs as those that cannot be compressed efficiently based on a data-driven graph encoding mechanism \cite{noble2003graph,nguyen2020anomaly}. 
One such mechanism is graph cover by motifs (or graphlets, substructures), wherein graphs that consist of frequent motifs can be encoded using relatively fewer bits, which is considered as the anomaly score.  
Note that motif-based encoding can be seen as a subgraph-based approach, discussed previously. The key distinction is that subgraph-based approaches define anomalies in terms of specific subgraphs (e.g. hotspots), whereas compression-based approaches quantify anomalies
implicitly in a data-driven fashion; a graph is deemed more anomalous the less it exhibits frequent motifs (here, rare motifs can be any complement, rather than specifically defined).

A key distinction of GAD from point-cloud OD is the abundance of anomaly definitions.
I tried to overview these in a two-level ``catalog'' here (e.g. node-level, feature-based). It would be interesting to identify other novel definitions and extend this existing catalog, ideally either based on a key driving application or otherwise with a strong justification for real-world applicability.
A key challenge in this area appears to be a general-purpose, unified GAD framework that can admit graphs with any (subset of) properties. Existing methods often tend to be strongly tied to graphs with certain properties, perhaps mainly because the anomalies they aim to detect are also defined based on the type of input graph they admit.
For example, an approach based on compressing node-labeled multi-graphs \cite{nguyen2020anomaly} or based on identifying dense-subgraphs in plain graphs \cite{beutel2013copycatch} 
 does not trivially apply to 
 multi-attributed nodes. 


\subsection{Representation Learning and Deep Graph Anomaly Detection}

As mentioned earlier, feature/similarity-based approaches to GAD is one of the most prevalent. To this end, a large body of work exists on graph feature extraction, graph similarity, and graph kernels. However, these are often confined to certain types of graphs they can admit and do not apply broadly (See e.g. Table 1 in \cite{kriege2020survey}).
 
The most recent trend is representation learning or graph embedding through graph neural networks (GNNs) \cite{hamilton2017representation}.
GNNs are appealing for GAD since they can automatically learn vector representations at node- and graph-level, can be trained end-to-end to do so provided suitable objective functions, and can
readily handle graphs with various properties such as labeled/multi-attributed nodes and directed, weighted edges.
As such, unsupervised embedding techniques provide a simple way for mapping a GAD problem to a point-cloud OD one.

In addition, there have been recent work on purposing GNNs specifically for GAD.
The main directions include end-to-end representation learning guided by auto-encoding or one-class classification \cite{yu2018netwalk,ding2019deep,DOroNMM19,wang2021one}, 
as well as
defining new loss functions suitable for GAD
\cite{zhao2020error}.

I have listed several advantages of GNNs above.
On the other hand, there are several challenges with effectively using GNNs for GAD.
First, end-to-end training necessitates suitable loss functions,
which limits the solution to a few existing loss function families for GAD.
Second, learned representations are hard to interpret, compared to e.g. hand-crafted features or motifs which may be more intuitive.
Third, GNNs may be more complicated and slow to train.
Last and perhaps most importantly, like their NN counterparts, they have a long list of (hyper)parameters. A big question is how to tune them in the absence of any labels.
There is recent work on leveraging meta-learning for effective graph anomaly detection assuming a few labels exist \cite{ding2021few}.
Future work on automating model selection could unlock the true potential of GNNs for GAD.


\section{Conclusions and Discussion}

In this article, I aimed to present a broad overview of the past, recently trending topics and potential future directions for both point-cloud outlier detection (OD) and graph anomaly detection (GAD). In the next couple of paragraphs, I give a short summary, followed by a discussion on open problems in anomaly mining at large.

Outlier definitions seem to have settled for point-cloud OD; as global, local/contextual, and collective outliers. Great progress has also been made in terms of designing detection models, for various settings such as streaming, high-dimensional, distributed, etc. data. Recent trends include different detection paradigms, such as fairness-aware, explainable, and human-in-the-loop detection.
Recent advances in deep neural networks (NNs) have also been carried over to this area for deep OD.
Specifically these recent deep NN models that are heavily (hyper)parameter (HP) laden, and more broadly the plethora of available detection models, have motivated automating OD, that is, both algorithm (i.e. detector) as well as HP selection using only a few labels or no labels at all.

On the other hand, and in contrast to point-cloud OD, anomaly definitions for GAD vary widely.
This is partly driven by the abundant different real-world applications to which GAD is relevant,
as well as the variety of additional properties a graph can be decorated with, such as edge multiplicities, node attributes, etc.
Moreover, while some definitions directly reflect the known types of anomalies (e.g., emergence or existence of dense subgraphs), others look for unknown anomalies as those outside the observed patterns (e.g. frequent motifs).
This implies that the way graph patterns are defined has direct implications for GAD. Moreover, the better pattern mining would imply better anomaly detection.
Deep learning, and specifically (G)NNs, is also a recent trend for GAD, for which suitable loss functions and effective model selection --esp. using no labels-- are key problems.



I conclude with discussing a couple of overarching open problems for anomaly mining.
The first is on \textbf{evaluation and benchmark datasets}.
It is most common practice for the community to re-purpose (point or graph) classification datasets, by down/sub-sampling one or more of the classes to depict the ``ground-truth'' outliers \cite{emmott2015meta}.
This avoids having to inject synthetic anomalies, whose generation may be prone to internal biases of how certain detection models work.
On the other hand, this procedure raises other concerns such as whether it creates suitable settings that align with what is being evaluated as well as human intuition, and whether it truly represents the nature of anomalies that arise in real world applications
\cite{Campos2016,swersky2016evaluation,zhao2020using}.
Certainly, rigorous attention to designing better benchmark datasets for evaluation 
would help reveal 
the
strengths and limitations of various detection models,  and ultimately keep a fair and accurate record of progress \cite{wu2020current}.
A promising way here is the industry putting out real world detection tasks, perhaps in the form of competitions with a leader-board, where the data may contain some identified (but perhaps not all) true anomalies. Understandably such data may be sensitive for public domain, where privacy-preserving data release practices could be employed. Specifically for OD, distance-preserving data sketches or a pairwise-distance oracle (instead of explicit feature representations) could enable ground for data exchange.

Another overarching open challenge is adaptive \textbf{detection under adversarial settings}.
Anomaly mining is used more often than not for applications involving  fraud (e.g. healthcare, advertisement, tax), intrusion, etc. where intelligent attackers continuously adapt their behavior to evade detection. There exist some work on detecting specific type of (dense subgraph injection) small-scale (called ``stealth'') and camouflaged anomalies \cite{shah2014spotting,hooi2016fraudar}, although these do not model the dynamic behavior of attackers. In domains such as banking and cybersecurity,  models in place are observed to stop detecting any anomalies soon after (within hours) of being deployed.
This calls for fast-adaptive or otherwise adversarially-robust detection algorithms.




\section*{Acknowledgments}
This work is sponsored by NSF CAREER 1452425. I also thank PwC Risk and Regulatory Services Innovation Center at Carnegie Mellon University. Any conclusions expressed in this material are those of the author and do not necessarily reflect the views, expressed or implied, of the funding parties.


\bibliographystyle{named}
\bibliography{ijcai21}

\end{document}